\documentclass{article}

\usepackage{graphicx} 
\usepackage{subfigure}

\usepackage{bm}  
\usepackage{amsmath}
\usepackage{amssymb}
\usepackage{amsfonts}
\usepackage{booktabs}
\usepackage{url}

\usepackage{graphicx}

\newtheorem{thm}{Theorem}[section]

\usepackage{natbib}

\usepackage{algorithm}
\usepackage{algorithmic}

\usepackage{hyperref}

\usepackage[accepted]{icml2013} 

\newcommand{\bmu}{\boldsymbol{\mu}}
\newcommand{\bK}{\boldsymbol{\mathrm{K}}}
\newcommand{\bM}{\boldsymbol{\mathrm{M}}}

\usepackage{color}

\icmltitlerunning{Gaussian Process Kernels for Pattern Discovery and Extrapolation}

\begin{document} 

\twocolumn[
\icmltitle{Gaussian Process Kernels for Pattern Discovery and Extrapolation}

\icmlauthor{Andrew Gordon Wilson}{agw38@cam.ac.uk}
\icmladdress{Department of Engineering, University of Cambridge, Cambridge, UK}
\icmlauthor{Ryan Prescott Adams}{rpa@seas.harvard.edu}
\icmladdress{School of Engineering and Applied Sciences, Harvard University, Cambridge, USA}

\icmlkeywords{Gaussian Processes, Bayesian Nonparametrics, Kernel Learning, Time Series, Feature Discovery, Extrapolation, Pattern Discovery, Gaussian Process, ICML}

\vskip 0.3in
]

\begin{abstract} 
Gaussian processes are rich distributions over functions, which
provide a Bayesian nonparametric approach to smoothing and
interpolation.  We introduce simple closed form kernels that can be
used with Gaussian processes to discover patterns and enable
extrapolation.  These kernels are derived by modelling a spectral
density -- the Fourier transform of a kernel -- with a
Gaussian mixture.  The proposed kernels support a broad class of stationary
covariances, but Gaussian process inference remains simple and
analytic.  We demonstrate the proposed kernels by discovering patterns
and performing long range extrapolation on synthetic examples, as well
as atmospheric CO$_2$ trends and airline passenger data.  We also show
that it is possible to reconstruct several popular standard covariances 
within our framework.
\end{abstract} 

\date{}

\section{Introduction}

Machine learning is fundamentally about pattern discovery.  The first
machine learning models, such as the perceptron
\citep{rosenblatt1962}, were based on a simple model of a neuron
\citep{mcculloch1943}.  Papers such as \citet{rumelhart1986learning}
inspired hope that it would be possible to develop intelligent agents
with models like neural networks, which could automatically discover
hidden representations in data.  Indeed, machine learning aims not
only to equip humans with tools to analyze data, but to fully automate
the learning and decision making process.

Research on Gaussian processes (GPs) within the machine learning
community developed out of neural networks research, triggered by
\citet{neal1996}, who observed that Bayesian neural networks became
Gaussian processes as the number of hidden units approached
infinity. \citet{neal1996} conjectured that ``there may be simpler
ways to do inference in this case''.

These simple inference techniques became the cornerstone of subsequent
Gaussian process models for machine learning \citep{rasmussen06}.
These models construct a prior directly over functions, rather than
parameters.  Assuming Gaussian noise, one can analytically infer a
posterior distribution over these functions, given data.  Gaussian
process models have become popular for non-linear regression and
classification \citep{rasmussen06}, and often have impressive
empirical performance \citep{rasmussenphd96}.

The properties of likely functions under a GP, e.g., smoothness,
periodicity, etc., are controlled by a positive definite
\emph{covariance kernel}\footnote{The terms \textit{covariance
kernel}, \textit{covariance function}, \textit{kernel function},
and \textit{kernel} are used interchangeably.}, an operator which
determines the similarity between pairs of points in the
domain of the random function.  The choice of kernel profoundly
affects the performance of a Gaussian process on a given task -- as
much as the choice of architecture, activation functions, learning
rate, etc., can affect the performance of a neural network.

Gaussian processes are sometimes used as expressive statistical tools,
where the pattern discovery is performed by a human, and then hard
coded into parametric kernels.  Often, however, the squared
exponential (Gaussian) kernel is used by default.  In either case,
GPs are used as smoothing interpolators with a fixed
(albeit infinite) set of basis functions.  Such simple smoothing
devices are not a realistic replacement for neural networks, which
were envisaged as intelligent agents that could discover hidden
features in data\footnote{In this paper, we refer to
  \textit{representations}, \textit{features} and \textit{patterns}
  interchangeably.  In other contexts, the term \textit{features}
  sometimes specifically means low dimensional representations of
  data, like neurons in a neural network.} via adaptive basis
functions \citep{mackay98}.

However, Bayesian nonparametrics can help build automated intelligent
systems that reason and make decisions.  It has been suggested that
the human ability for inductive reasoning -- concept generalization
with remarkably few examples -- could derive from a prior combined
with Bayesian inference \citep{yuille2006vision,tenenbaum2011,
  steyvers2006}.  Bayesian nonparametric models, and Gaussian
processes in particular, are an expressive way to encode prior
knowledge, and also reflect the belief that the real world is
infinitely complex \citep{neal1996}.

With more expressive kernels, one could use Gaussian processes to
learn hidden representations in data.  Expressive kernels have been
developed by combining Gaussian processes in a type of Bayesian neural
network structure \citep{salakhutdinov2008, wilson12icml,
  damianou2012deep}.  However, these approaches, while promising,
typically 1)~are designed to model specific types of structure (e.g.,
input-dependent correlations between different tasks); 2)~make use of
component GPs with simple interpolating kernels;
3)~indirectly induce complicated kernels that do not have a closed form
and are difficult to interpret; and 4)~require sophisticated
approximate inference techniques that are much more demanding than
that required by simple analytic kernels.

Sophisticated kernels are most often achieved by composing together a
few standard kernel functions
\citep{archambeau2011,durrande2011,gonen2011,rasmussen06}.  Tight
restrictions are typically enforced on these compositions and they are
hand-crafted for specialized applications.  Without such restrictions,
complicated compositions of kernels can lead to overfitting and
unmanageable hyperparameter inference.  Moreover, while some
compositions (e.g., addition) have an interpretable effect, 
many other operations change the
distribution over functions in ways that are difficult to identify.
It is difficult, therefore, to construct an effective inductive bias
for kernel composition that leads to automatic discovery of the
appropriate statistical structure, without human intervention.

This difficulty is exacerbated by the fact that it is challenging to
say anything about the covariance function of a stochastic process
from a single draw if \textit{no} assumptions are made.  If we allow
the covariance between any two points in the input space to arise from
\emph{any} positive definite function, with equal probability, then we gain 
essentially no information from a single realization.  Most commonly one assumes a
restriction to \emph{stationary} kernels, meaning that covariances
are invariant to translations in the input space.

In this paper, we explore flexible classes of kernels that go beyond
composition of simple analytic forms, while maintaining the useful
inductive bias of stationarity.  We propose new kernels which can be
used to automatically discover patterns and extrapolate far beyond the
available data.  This class of kernels contains many stationary
kernels, but has a simple closed form that leads to straightforward
analytic inference.  The simplicity of these kernels is one of their
strongest qualities.  In many cases, these kernels can be used as a
drop in replacement for the popular squared exponential kernel, with
benefits in performance and expressiveness.  By learning features in
data, we not only improve predictive performance, but we can more
deeply understand the structure of the problem at hand -- greenhouse
gases, air travel, heart physiology, brain activity, etc.

After a brief review of Gaussian processes in Section \ref{sec: gps},
we derive the new kernels in Section \ref{sec: easycov} by modelling a
spectral density with a mixture of Gaussians.  We focus our
experiments in Section \ref{sec: experiments} on elucidating the
fundamental differences between the proposed kernels and the popular
alternatives in \citet{rasmussen06}.  In particular, we show how the
proposed kernels can automatically discover patterns and extrapolate
on the CO$_2$ dataset in \citet{rasmussen06}, on a synthetic dataset
with strong negative covariances, on a difficult synthetic sinc
pattern, and on airline passenger data.  We also use our framework
to reconstruct several popular standard kernels.

\section{Gaussian Processes}
\label{sec: gps}

A Gaussian process is a collection of random variables, any finite
number of which have a joint Gaussian distribution.  Using a Gaussian
process, we can define a distribution over functions $f(x)$,
\begin{equation}
 f(x) \sim \mathcal{GP}(m(x),k(x,x')) \,,
\end{equation}
where $x \in \mathbb{R}^{P}$ is an arbitrary input variable, and the
mean function $m(x)$ and covariance kernel $k(x,x')$ are defined as
\begin{align}
 m(x) &= \mathbb{E}[f(x)] \,, \\
 k(x,x') &= \text{cov}(f(x),f(x')) \,. \label{eqn: covgeneral}
\end{align}
Any collection of function values has a joint Gaussian distribution
\begin{align}
 [f(x_1), f(x_2), \dots, f(x_N)]^{\top} &\sim \mathcal{N}(\bmu,\bK) \,,
\end{align}
where the ${N \times N}$ covariance matrix~$\bK$ has entries~${\mathrm{K}_{ij} =
k(x_i,x_j)}$, and the mean $\bmu$ has entries~${\mu_i =
m(x_i)}$.  The properties of the functions -- smoothness, periodicity,
etc. -- are determined by the kernel function.

The popular \textit{squared exponential} (SE) kernel has the form 
\begin{equation}
 k_{\text{SE}}(x,x') = \text{exp}(-0.5 ||x-x'||^2/\ell^2) \,. \label{eqn: sekernel}
\end{equation}
Functions drawn from a Gaussian process with this kernel function are
infinitely differentiable, and can display long range trends.
GPs with a squared exponential kernel are simply
smoothing devices: the only covariance structure that can be learned
from data is the length-scale $\ell$, which determines how quickly a
Gaussian process function varies with $x$.

Assuming Gaussian noise, one can analytically infer a posterior
predictive distribution over Gaussian process functions, and
analytically derive a marginal likelihood of the observed function
values $\bm{y}$ given only hyperparameters $\theta$, and the input
locations~$\{x_n\}^N_{n=1}$, $p(\bm{y}|\theta,\{x_n\}^N_{n=1})$.  This
marginal likelihood can be optimised to estimate 
hyperparameters such as~$\ell$, or used to integrate out the
hyperparameters via Markov chain Monte Carlo
\citep{murray-adams-2010a}.  Detailed Gaussian process 
references include \citet{rasmussen06}, \citet{stein1999},
and \citet{cressie1993statistics}.

\section{Kernels for Pattern Discovery}
\label{sec: easycov}
In this section we introduce a class of kernels that can discover
patterns, extrapolate, and model negative covariances.  This class
contains a large set of stationary kernels.  Roughly, a kernel
measures the similarity between data points.  As in Equation
\eqref{eqn: covgeneral}, the covariance kernel of a GP determines how
the associated random functions will tend to vary with inputs
(predictors) $x \in \mathbb{R}^P$.  A \textit{stationary kernel} is a
function of $\tau = x-x'$, i.e., it is invariant to translation of the
inputs.

Any stationary kernel (aka covariance function) can be expressed as an
integral using Bochner's theorem \citep{bochner1959lectures,stein1999}:
\begin{thm}
 (Bochner) A complex-valued function k on $\mathbb{R}^{P}$ is the
  covariance function of a weakly stationary mean square continuous
  complex-valued random process on $\mathbb{R}^P$ if and only if it
  can be represented as
\begin{equation}
 k(\tau) = \int_{\mathbb{R}^{P}} e^{2\pi i s^{\top} \tau} \psi(\mathrm{d}s) \,,
\end{equation}
where $\psi$ is a positive finite measure.
\end{thm}
If $\psi$ has a density $S(s)$, then $S$ is called the 
\textit{spectral density} or \textit{power spectrum} of
$k$, and $k$ and $S$ are Fourier duals \citep{chatfield1989}:
\begin{align}
 k(\tau) &= \int S(s)e^{2\pi i s^{\top}\tau} ds \,, \label{eqn: kernelfourier} \\
 S(s) &= \int k(\tau)e^{-2\pi i s^{\top} \tau} d\tau \label{eqn: spectralfourier} \,.
\end{align}

In other words, a spectral density entirely determines the properties
of a stationary kernel.  Substituting the squared exponential kernel
of \eqref{eqn: sekernel} into \eqref{eqn: spectralfourier}, we find
its spectral density is $S_{\text{SE}}(s) = (2\pi \ell^2)^{P/2}
\exp(-2\pi^2\ell^2s^2)$.  Therefore SE kernels, and mixtures of SE
kernels, are a very small corner of the set of possible stationary
kernels, as they correspond only to Gaussian spectral densities
centered on the origin.

However, by using a mixture of Gaussians that have non-zero means, one
can achieve a much wider range of spectral densities.  Indeed,
mixtures of Gaussians are dense in the set of all distribution
functions \citep{kostantinos2000}.  Therefore, the dual of this set is also
dense in stationary covariances.  That is, we can approximate
\emph{any} stationary covariance kernel to arbitrary precision, given
enough mixture components in the spectral representation.  This
observation motivates our approach, which is to model GP
covariance functions via spectral densities that are scale-location
mixtures of Gaussians.

We first consider a simple case, where
\begin{align}
\phi(s\,;\mu,\sigma^2) &=
\frac{1}{\sqrt{2 \pi \sigma^2}} \exp\{-\frac{1}{2\sigma^2}(s-\mu)^2\},\quad\text{and}\\
S(s) & = [\phi(s) + \phi(-s)]/2 \,,
\end{align}
noting that spectral densities are symmetric \citep{rasmussen06}.
Substituting $S(s)$ into equation \eqref{eqn: kernelfourier}, we find
\begin{align}
k(\tau) &= \exp\{-2 \pi^2 \tau^2 \sigma^2 \} \cos(2\pi \tau \mu) \,.  \label{eqn: smcomponentkernel}
\end{align}
If~$\phi(s)$ is instead a mixture of~$Q$ Gaussians on~$\mathbb{R}^P$,
where the~$q^\text{th}$ component has mean vector~${\bm{\mu}_q =
  (\mu_q^{(1)},\dots,\mu_q^{(P)})}$
and covariance matrix~${\bM_q = \text{diag}(v_q^{(1)},\dots,v_q^{(P)})}$,
and~$\tau_p$ is the~$p^{\text{th}}$ component of the~$P$ dimensional
vector~${\tau=x-x'}$, then
\begin{align}
 k(\tau) &= \sum_{q=1}^{Q} \!w_q\! \prod_{p=1}^{P} \exp\{-2\pi^2\tau_p^2v_q^{(p)}\}\cos(2\pi \tau_p \mu_q^{(p)}).  \label{eqn: smkernel}
\end{align}
The integral in \eqref{eqn: kernelfourier} is tractable even when the
spectral density is an arbitrary Gaussian mixture, allowing us to
derive\footnote{Detailed derivations of Eqs.~\eqref{eqn: smcomponentkernel}
  and \eqref{eqn: smkernel} are in the supplementary material \citep{wilsonsuppnew}.} the
exact closed form expressions in Eqs.~\eqref{eqn: smcomponentkernel} and
\eqref{eqn: smkernel}, and to perform analytic inference with Gaussian
processes. Moreover, this class of kernels is expressive -- containing
many stationary kernels -- but nevertheless has a simple form.

These kernels are easy to interpret, and provide drop-in replacements
for kernels in \citet{rasmussen06}.  The weights~$w_q$ specify the
relative contribution of each mixture component.  The inverse
means~${1/\mu_q}$ are the component periods, and the inverse standard
deviations~${1/\sqrt{v_q}}$ are length-scales, determining how quickly
a component varies with the inputs~$x$.  The kernel in Eq.~\eqref{eqn:
  smkernel} can also be interpreted through its associated spectral
density.  In Section \ref{sec: experiments}, we use the learned
spectral density to interpret the number of discovered patterns in the
data and how these patterns generalize.  Henceforth, we refer to the
kernel in Eq.~\eqref{eqn: smkernel} as a spectral mixture (SM) kernel.

\section{Experiments}
\label{sec: experiments} 

We show how the SM kernel in Eq.~\eqref{eqn: smkernel} can be used to
discover patterns, extrapolate, and model negative covariances.  We
contrast the SM kernel with popular kernels in, e.g., \citet{rasmussen06}
and \citet{abrahamsen1997},
which typically only provide smooth interpolation.  Although the SM
kernel generally improves predictive likelihood over popular
alternatives, we focus on clearly visualizing the learned kernels and
spectral densities, examining patterns and predictions, and
discovering structure in the data.  Our objective is to elucidate the
fundamental differences between the proposed SM kernel and the
alternatives. 

In all experiments, Gaussian noise is assumed, so that marginalization
over the unknown function can be performed in closed form.  Kernel
hyperparameters are trained using nonlinear conjugate gradients to
optimize the marginal likelihood~$p(\bm{y}|\theta,\{x_n\}^N_{n=1})$ of
the data $\bm{y}$ given hyperparameters $\theta$, as described in
Section \ref{sec: gps}, assuming a zero mean GP.  A type of ``automatic relevance
determination'' \citep{mackay1994,tipping2004} takes place during
training, minimizing the effect of extraneous components in the proposed model,
through the complexity penalty in the marginal likelihood \citep{rasmussen06}.
The complexity penalty in the log marginal likelihood is a log determinant which
can be written as a sum of eigenvalues of a covariance matrix.  Extraneous weights
shrink towards zero, since they will increase the eigenvalues of this covariance 
matrix, without significantly improving the model fit.  Moreover, the exponential terms in the SM kernel of Eq.~\eqref{eqn: smkernel}
have an annealing effect on the marginal likelihood, reducing 
multimodality in the frequency parameters, making it easier to 
naively optimize the marginal likelihood 
without converging to undesirable local optima.  For
a fully Bayesian treatment, the spectral density could alternatively
be integrated out using Markov chain Monte Carlo
\citep{murray-adams-2010a}, rather than choosing a point estimate.
However, we wish to emphasize that the SM kernel can be successfully
used in the same way as other popular kernels, without additional
inference efforts. 

We compare with the popular squared exponential (SE), 
Mat{\'e}rn (MA), rational quadratic (RQ), and 
periodic (PE) kernels.  In each comparison, we attempt to give these alternative kernels fair
treatment: we initialise hyperparameters at values that give high
marginal likelihoods and which are well suited to the datasets, based
on features we can already see in the data.  Conversely, we randomly
initialise the parameters for the SM kernel.  Training runtimes are on the order of minutes
for all tested kernels.  In these examples, comparing with multiple kernel learning (MKL)
\citep{gonen2011} has limited additional value.  MKL is not typically intended for 
pattern discovery, and often uses mixtures of SE kernels.  Mixtures of SE kernels
correspond to scale-mixtures of Gaussian spectral densities, and do not perform well on these data, 
which are described by highly multimodal non-Gaussian spectral densities.

\subsection{Extrapolating Atmospheric CO$_2$}
\label{sec: CO2}

\begin{figure}
\centering%
\includegraphics[scale=.47]{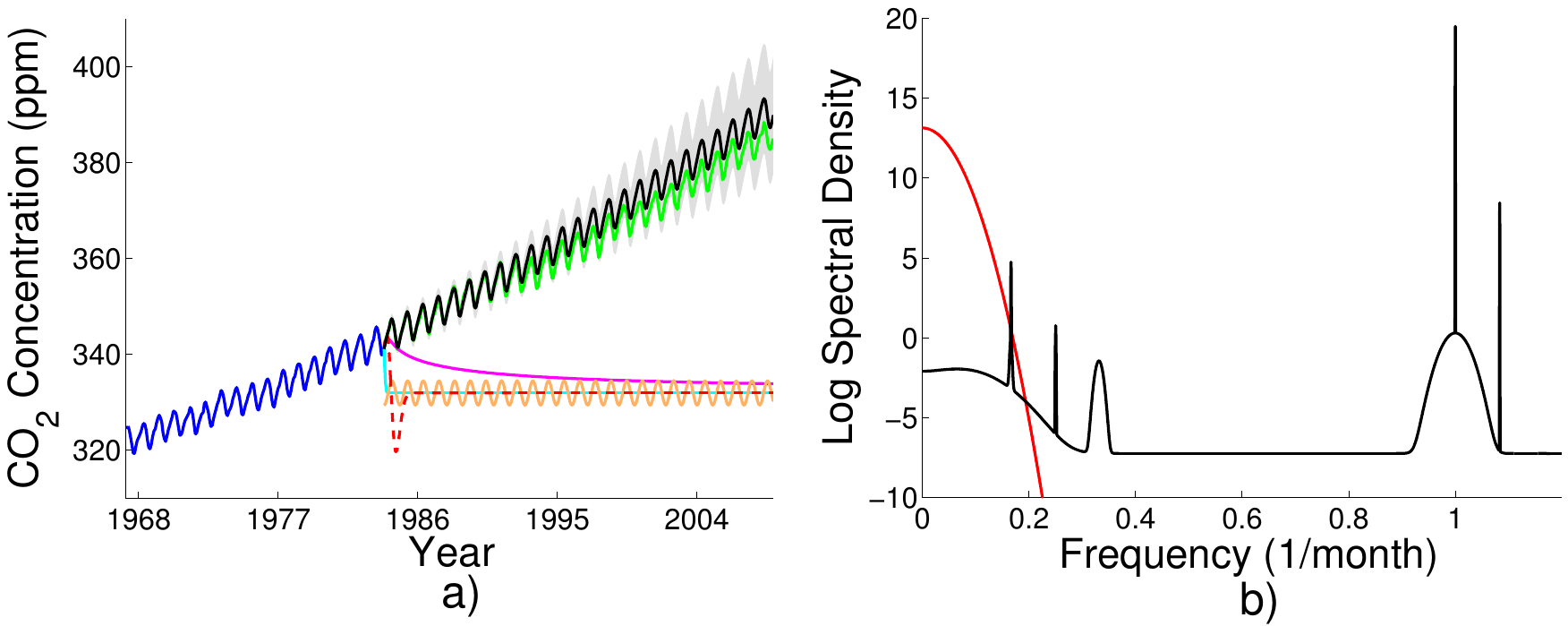}
\caption{\small Mauna Loa CO$_2$ Concentrations.  a) Forecasting
  CO$_2$.  The training data are in blue, the testing data in green.
  Mean forecasts made using the SM kernel are in black, with 2 stdev
  about the mean (95\% of the predictive mass) in gray shade.
  Predictions using the Mat{\'e}rn (MA), squared exponential (SE),
  rational quadratic (RQ), and periodic kernels (PE) are in cyan, dashed red,
  magenta, and orange, respectively.  b) The log spectral densities of
  the learned SM and squared exponential kernels are in black and red,
  respectively.}
\label{fig: CO2recon}
\end{figure}

\citet{keeling2004} recorded monthly average atmospheric CO$_2$
concentrations at the Mauna Loa Observatory, Hawaii.  The data are
shown in Figure \ref{fig: CO2recon}.  The first $200$ months are used
for training (in blue), and the remaining $301$ months ($\approx 25$ years) are used for
testing (in green).

This dataset was used in \citet{rasmussen06}, and is frequently used
in Gaussian process tutorials, to show how GPs are flexible
statistical tools: a human can look at the data, recognize patterns,
and then hard code those patterns into covariance kernels.
\citet{rasmussen06} identify, by looking at the blue and green curves
in Figure \ref{fig: CO2recon}a, a long term rising trend, seasonal
variation with possible decay away from periodicity, medium term
irregularities, and noise, and hard code a stationary covariance
kernel to represent each of these features.

However, in this view of GP modelling, all of the interesting pattern
discovery is done by the human user, and the GP is used
simply as a smoothing device, with the flexibility to incorporate
human inferences in the prior.  Our contention is that such pattern
recognition can also be performed algorithmically.  To discover these
patterns without encoding them \emph{a priori} into the GP, we use the
spectral mixture kernel in Eq.~\eqref{eqn: smkernel},
with~${Q=10}$.  The results are shown in Figure \ref{fig: CO2recon}a.

In the training region, predictions using each kernel are essentially
equivalent, and entirely overlap with the training data.  However,
unlike the other kernels, the SM kernel (in black) is able to discover
patterns in the training data and accurately extrapolate over a long
range.  The 95\% high predictive density (HPD) region contains the
true CO$_2$ measurements for the duration of the measurements.

We can see the structure discovered by the SM kernel in the learned
log spectral density in Figure \ref{fig: CO2recon}b.  Of the~${Q=10}$
components, only seven were used -- an example of automatic relevance
determination during training.  Figure
\ref{fig: CO2recon}b is a good example of \textit{aliasing} in the 
spectral density.  The sharp peak at a frequency near $1.08$ corresponds
to the period of $12$ months, relating to pronounced yearly variations in 
CO$_2$ readings.  Since $1.08$ is greater than the sampling rate of the
data, 1/month, it will be aliased back to $0.08$, and $1/0.08 = 12$ months.
Since there is little width to the peak, the model is confident that this 
feature (yearly periodicity) should extrapolate long into the future.
There is another large peak at a frequency of $1$, equal to the sampling
rate of the data.  This peak corresponds to an aliasing of a mean function,
and can also be related to noise aliasing.  In these examples, aliasing will not 
hurt predictions, but can affect the interpretability of results.
It is easy to stop aliasing, for example, by restricting the learned frequencies 
to be less than the Nyquist frequency (1/2 of the sampling rate of the data).
For pedagogical reasons -- and since it does not affect performance in these
examples -- we have
not used such restrictions here.  Finally, we see other 
peaks corresponding to periods of 6 months, 4 months, 3 months and 1 month. 
The width of each peak reflects the model's uncertainty about the 
corresponding feature in the data.  In red, we show the spectral density for the learned SE kernel, which
misses many of the frequencies identified as important using the SM 
kernel.  

All SE kernels have a Gaussian spectral density
centred on zero.  Since a mixture of Gaussians centred on the origin is a
poor approximation to many density functions,
combinations of SE kernels have limited expressive power.  Indeed the
spectral density learned by the SM kernel in Figure \ref{fig:
  CO2recon}b is highly non-Gaussian.  The test predictive performance
using the SM, SE, MA, RQ, and PE kernels is given in Table \ref{tab:
  predictions}, under the heading \texttt{CO}$_\texttt{2}$.
\subsection{Recovering Popular Kernels}
\label{sec: recovering}

\begin{figure}
\centering
\includegraphics[scale=.47]{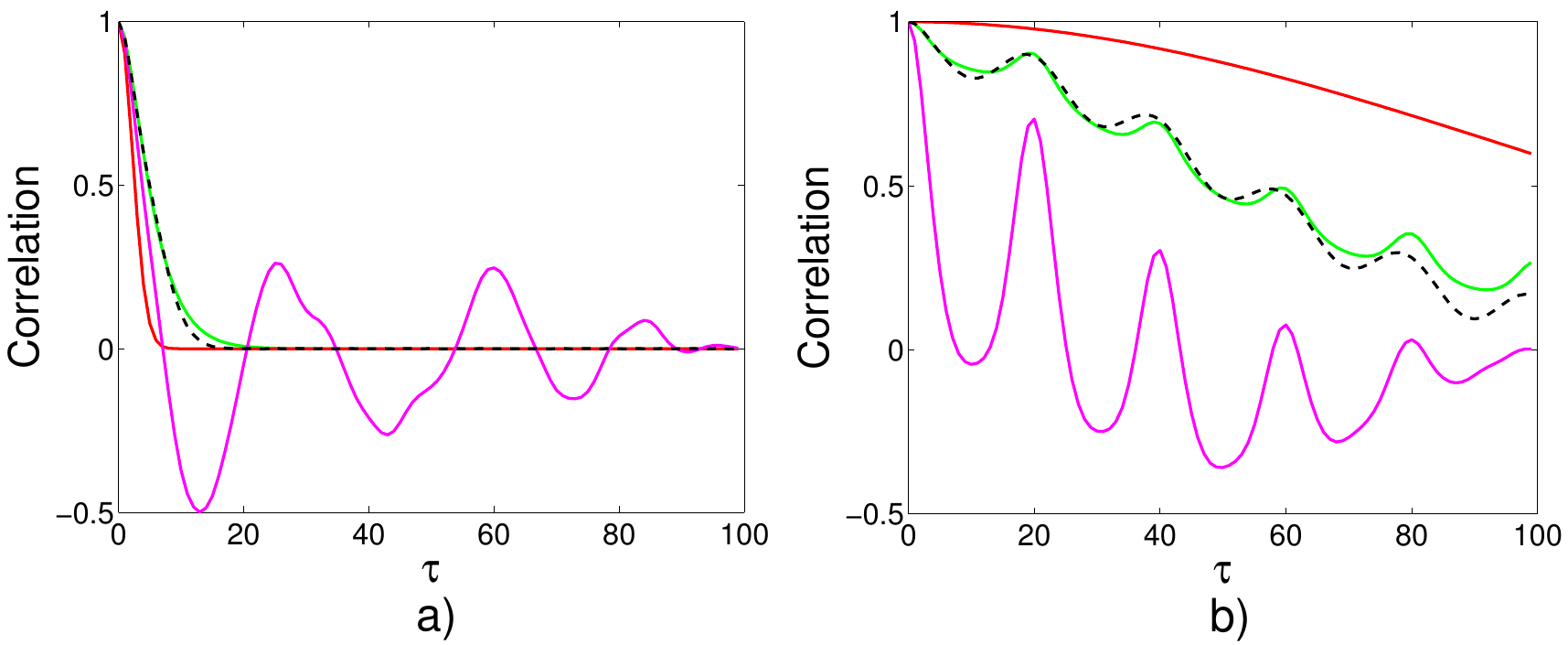}
\caption{\small Recovering popular correlation functions (normalised
  kernels), with~${\tau = x-x'}$.  The true correlation function
  underlying the data is in green, and SM, SE, and empirical
  correlation functions are in dashed black, red, and magenta,
  respectively.  Data are generated from a) a Mat{\'e}rn kernel, and
  b) a sum of RQ and periodic kernels.}
\label{fig: covlearn}
\end{figure}

The SM class of kernels contains many stationary kernels, since
mixtures of Gaussians can be used to construct a wide range of
spectral densities.  Even with a small number of mixture
components, e.g.,~${Q \leq 10}$, the SM kernel can closely recover 
popular stationary kernels catalogued in \citet{rasmussen06}.

As an example, we start by sampling $100$ points from a
one-dimensional GP with a Mat{\'e}rn kernel with degrees
of freedom ~${\nu = 3/2}$:
\begin{equation}
k_{\text{MA}}(\tau) =
a(1+\frac{\sqrt{3}\tau}{\ell})\exp(-\frac{\sqrt{3}\tau}{\ell}) \,,
\end{equation}
where ${\ell = 5}$ and~${a=4}$.  Sample functions from a Gaussian
process with this Mat{\'e}rn kernel are far less smooth (only
once-differentiable) than Gaussian process functions with a squared
exponential kernel.
                                                                                              
We attempt to reconstruct the kernel underlying the data by training
an SM kernel with~${Q=10}$.  After training, only~${Q=4}$
components are used.  The log marginal likelihood of the data --
having integrated away the Gaussian process -- using the trained SM
kernel is $-133$, compared to $-138$ for the Mat{\'e}rn kernel that
generated the data.  Training the SE kernel in
\eqref{eqn: sekernel} gives a log marginal likelihood of $-140$.

Figure \ref{fig: covlearn}a shows the learned SM correlation
function\footnote{A \textit{correlation function}~$c(x,x')$ is a
  normalised covariance kernel~$k(x,x')$, such that~${c(x,x') =
    k(x,x')/\sqrt{k(x,x)k(x',x')}}$ and~${c(x,x)=1}$.}, compared to
the generating Mat{\'e}rn correlation function, the empirical
autocorrelation function, and learned squared exponential correlation
function.  Although often used in geostatistics to guide choices of
GP kernels (and parameters)
\citep{cressie1993statistics}, the empirical autocorrelation function
tends to be unreliable, particularly with a small amount of data
(e.g.,~${N < 1000}$), and at high lags (for~${\tau > 10}$).  In Figure
\ref{fig: covlearn}a, the empirical autocorrelation function is erratic
and does not resemble the Mat{\'e}rn kernel for~${\tau > 10}$. Moreover,
the squared exponential kernel cannot capture the heavy tails of the
Mat{\'e}rn kernel, no matter what length-scale it has.   Even though the SM 
kernel is infinitely differentiable, it can closely approximate processes
which are finitely differentiable, because mixtures of Gaussians can closely 
approximate the spectral densities of these processes, even 
with a small number of components, as in Figure \ref{fig: covlearn}a.

Next, we reconstruct a mixture of the rational quadratic (RQ) and periodic kernels (PE) in \citet{rasmussen06}:
\begin{align}
k_{\text{RQ}}(\tau) &= (1+\frac{\tau^2}{2\,\alpha \,\ell_{RQ}^2})^{-\alpha} \,,\label{eqn: rqk}  \\ 
k_{\text{PE}}(\tau) &=  \exp(-2\sin^2(\pi\,\tau\,\omega)/\ell_{PE}^2)  \,. \label{eqn: pek}
\end{align}

The rational quadratic kernel in \eqref{eqn: rqk} is derived as a
scale mixture of squared exponential kernels with different
length-scales.  The standard periodic kernel in \eqref{eqn: pek} is
derived by mapping the two dimensional variable~${u(x) =
  (\cos(x),\sin(x))}$ through the squared exponential kernel in
\eqref{eqn: sekernel}.  Derivations for both the RQ and PE kernels in
\eqref{eqn: rqk} and \eqref{eqn: pek} are in \citet{rasmussen06}.
Rational quadratic and Mat{\'e}rn kernels are also discussed in
\citet{abrahamsen1997}.  We sample $100$ points from a Gaussian
process with kernel~${10 k_{\text{RQ}} + 4 k_{\text{PE}}}$, with~${\alpha
= 2}$,~${\omega = 1/20}$,~${\ell_{RQ} = 40}$,~${\ell_{PE} = 1}$.

We reconstruct the kernel function of the sampled process using an SM
kernel with~${Q=4}$, with the results shown in Figure \ref{fig:
  covlearn}b.  The heavy tails of the RQ kernel are modelled by two
components with large periods (essentially aperiodic), and one short
length-scale and one large length-scale. The third component has a relatively short
length-scale, and a period of~$20$.  There is not enough information
in the~$100$ sample points to justify using more than~${Q=3}$
components, and so the fourth component in the SM kernel has no effect,
through the complexity penalty in the marginal likelihood.
The empirical autocorrelation function somewhat captures the
periodicity in the data, but significantly underestimates the
correlations.  The squared exponential kernel learns a long
length-scale: since the SE kernel is highly misspecified with
the true generating kernel, the data are explained as noise.

\subsection{Negative Covariances}
\label{sec: negativecov}
\begin{figure}
\centering
\includegraphics[scale=.57]{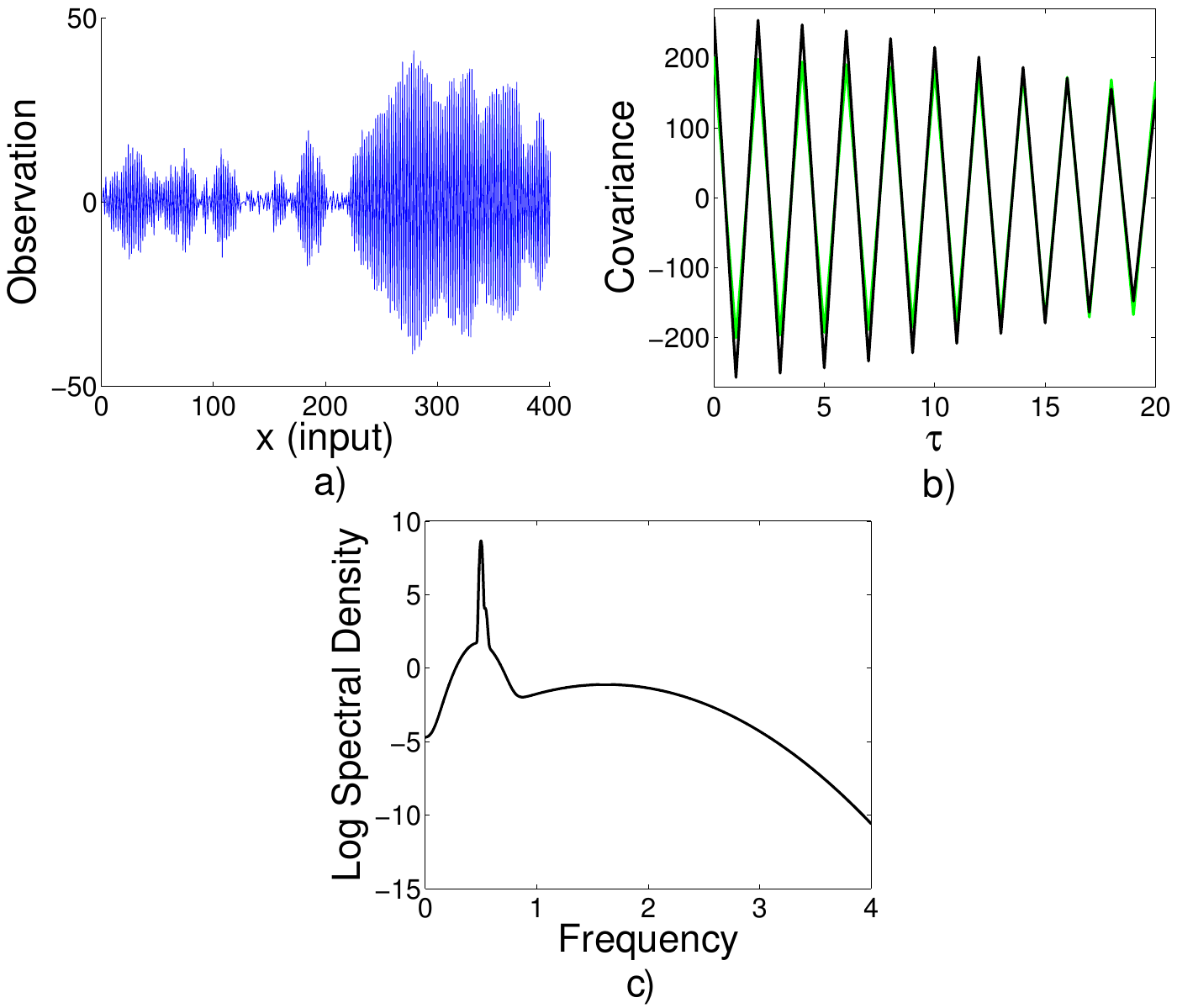}
\caption{\small Negative Covariances.  a) Observations
of a discrete time autoregressive (AR) series with negative
covariances.  b) The SM learned kernel is shown in black,
while the true kernel of the AR series is in green, with
$\tau = x-x'$.  
c) The spectral density of the learned SM kernel is
in black.}
\label{fig: covnegfigures}
\end{figure}

All of the stationary covariance functions in the standard machine
learning Gaussian process reference \citet{rasmussen06} are everywhere
positive, including the periodic kernel,~${k(\tau) =
\exp(-2\sin^2(\pi\,\tau\,\omega)/\ell^2)}$.  While positive covariances are
often suitable for interpolation, capturing negative covariances can
be essential for extrapolating patterns: for example, linear trends
have long-range negative covariances.  We test the ability of the SM
kernel to learn negative covariances, by sampling 400 points
from a simple AR(1) discrete time GP:
\begin{align}
y(x+1) &= - e^{-0.01} y(x) + \sigma \epsilon(x) \,, \label{eqn: arprocess} \\
\epsilon(x) &\sim \mathcal{N}(0,1) \,, 
\end{align}
which has kernel
\begin{equation}
k(x,x') = \sigma^2 (-e^{-.01})^{|x-x'|}/(1-e^{-.02}) \,. \\
\end{equation}
The process in Eq.~\eqref{eqn: arprocess} is shown in Figure \ref{fig:
  covnegfigures}a.  This process follows an oscillatory pattern,
systematically switching states every~${x=1}$ unit, but is not
periodic and has long range covariances: if we were to only view every
second data point, the resulting process would vary rather slowly and
smoothly.

We see in Figure \ref{fig: covnegfigures}b that the learned SM
covariance function accurately reconstructs the true covariance
function.  The spectral density in Figure \ref{fig: covnegfigures}c)
shows a sharp peak at a frequency of $0.5$, or a period of $2$.  This
feature represents the tendency for the process to oscillate from
positive to negative every $x=1$ unit.  In this case $Q=4$, but after
automatic relevance determination in training, only 3 components were
used. Using the SM kernel, we forecast 20 units ahead and compare to
other kernels in Table \ref{tab: predictions} (\texttt{NEG COV}).

\subsection{Discovering the Sinc Pattern}
\label{sec: sinc}

\begin{figure}
\centering
\includegraphics[scale=.57]{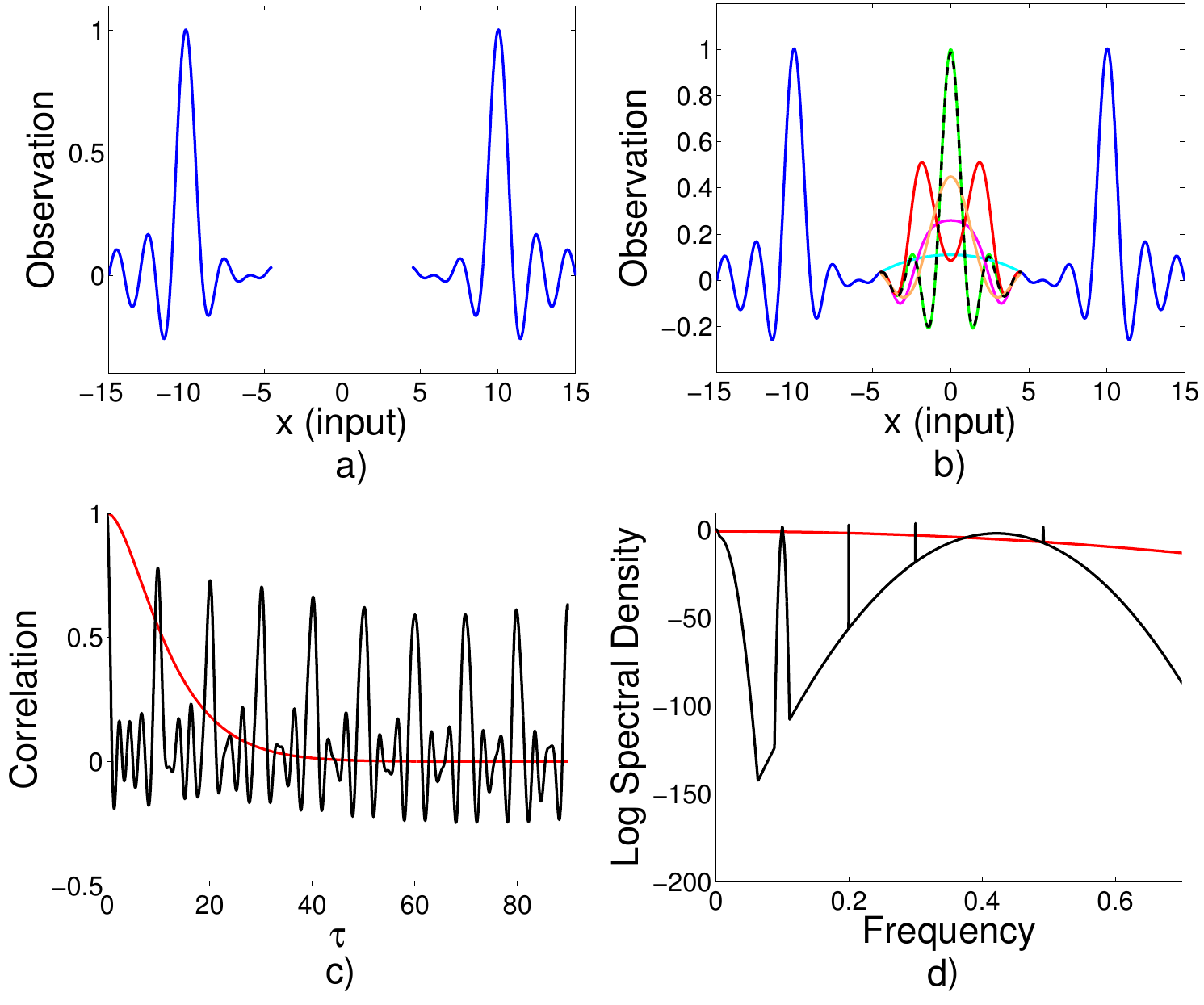}
\caption{\small Discovering a sinc pattern. a) The training
data are shown in blue.  The goal is to fill in the missing
region $x \in [-4.5,4.5]$.  b) The training data are in blue,
the testing data in green.  The mean of the predictive 
distribution using the SM kernel is shown in dashed
black. The mean of the predictive distributions using the 
squared exponential, Mat{\'e}rn, rational quadratic, 
and periodic kernels, are 
in red, cyan, magenta, and orange. c) The learned 
SM correlation function (normalised kernel) is shown 
in black, and the learned Mat{\'e}rn correlation function
is in red, with $\tau = x-x'$.  d)  The log spectral densities of the 
SM and SE kernels are in black and red,
respectively.}
\label{fig: sincfigures}
\end{figure}

The sinc function is defined as~${\text{sinc}(x) = \sin(\pi x)/(\pi
x)}$.  We created a pattern combining three sinc functions:
\begin{align}
 y(x) &= \text{sinc}(x+10) + \text{sinc}(x) + \text{sinc}(x-10) \,.
\end{align}
This is a complex oscillatory pattern.  Given only the points shown in
Figure \ref{fig: sincfigures}a, we wish to complete the pattern
for~${x \in [-4.5,4.5]}$.  Unlike the CO$_2$ example in Section
\ref{sec: CO2}, it is perhaps even difficult for a human to
extrapolate the missing pattern 
from the training data.  It is an interesting exercise to focus on this
figure, identify features, and fill in the missing part.

Notice that there is complete symmetry about the origin~$x=0$, peaks
at~${x=-10}$ and~${x=10}$, and destructive interference on each side of
the peaks facing the origin.  We therefore might
expect a peak at~${x=0}$ and a symmetric pattern around~${x=0}$.

As shown in Figure \ref{fig: sincfigures}b, the SM kernel
with~${Q=10}$ reconstructs the pattern in the region~${x \in
  [-4.5,4.5]}$ almost perfectly from the~$700$ training points in
blue.  Moreover, using the SM kernel, 95\% of the posterior predictive
mass entirely contains the true pattern in~${x \in [-4.5,4.5]}$.  GPs
using Mat{\'e}rn, SE, RQ, and
periodic kernels are able to predict reasonably within~${x=0.5}$ units
of the training data, but entirely miss the pattern in~${x \in
[-4.5,4.5]}$.

Figure \ref{fig: sincfigures}c shows the learned SM correlation
function (normalised kernel).  For ~${\tau \in [0,10]}$ there is a local
pattern, roughly representing the behaviour of a single~$\text{sinc}$
function. For~${\tau > 10}$ there is a repetitive pattern representing a
new sinc function every 10 units -- an extrapolation a human might
make.  With a sinc functions centred at~${x=-10, 0, 10}$, we might
expect more sinc patterns every $10$ units.  The learned Mat{\'e}rn
correlation function is shown in red in Figure \ref{fig:
  sincfigures}c -- unable to discover complex patterns in the data,
it simply assigns high correlation to nearby points.

Figure \ref{fig: sincfigures}d shows the (highly non-Gaussian)
spectral density of the SM kernel, with peaks at ${0.003, 0.1, 0.2,
0.3, 0.415, 0.424, 0.492}$.  In this case, only~$7$ of the~${Q=10}$
components are used.  The peak at~$0.1$ represents a period of~$10$:
every 10 units, a sinc function is repeated.  The variance of this
peak is small, meaning the method will extrapolate this structure over
long distances.  By contrast, the squared exponential spectral density
simply has a broad peak, centred at the origin.  The predictive
performance for recovering the missing~$300$ test points (in green) is
given in Table \ref{tab: predictions} (\texttt{SINC}).

\subsection{Airline Passenger Data}
\label{sec: airline}

\begin{figure}
\centering
\includegraphics[scale=.47]{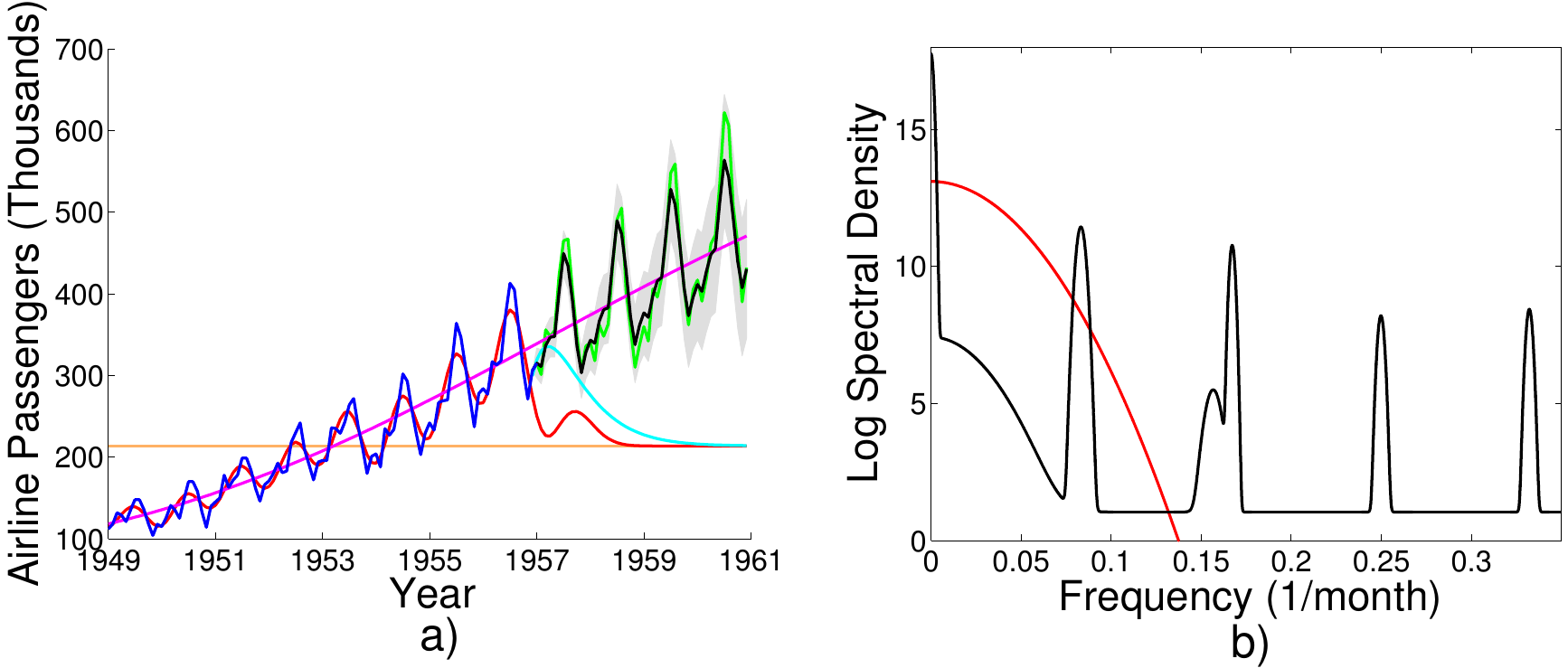}
\caption{\small Predicting airline passenger
numbers. a) The training and testing data
are in blue and green, respectively.  The mean of 
the predictive distribution using the SM kernel is 
shown in black, with 2 standard deviations about 
the mean (95\% of the predictive probability mass) 
shown in gray shade. The mean of the predictive 
distribution using the SM kernel is shown in 
black. The mean of the predictive distributions using the 
squared exponential, Mat{\'e}rn, rational quadratic,  
and periodic kernels, are in red, cyan, magenta, and 
orange, respectively.  In the training
region, the SM and Mat{\'e}rn kernels are not shown, since
their predictions essentially overlap with the 
training data.
b) The log spectral densities of the 
SM and squared exponential kernels are in black and red,
respectively.}
\label{fig: airlinefigures}
\end{figure}

Figure \ref{fig: airlinefigures}a shows airline passenger numbers,
recorded monthly, from 1949 to 1961 \citep{hyndman2005}.  Based on
only the first 96 monthly measurements, in blue, we wish to forecast
airline passenger numbers for the next 48 months (4 years); the
corresponding 48 test measurements are in green.  

There are several features apparent in these data: short seasonal
variations, a long term rising trend, and an absence of white noise
artifacts.  Many popular kernels are forced to make one of two
choices: 1)~Model the short term variations and ignore the long term
trend, at the expense of extrapolation.  2)~Model the long term trend
and treat the shorter variations as noise, at the expense of
interpolation.

As seen in Figure \ref{fig: airlinefigures}a, the Mat{\'e}rn kernel
is more inclined to model the short term trends than the smoother SE
or RQ kernels, resulting in sensible interpolation (predicting almost
identical values to the training data in the training region), but
poor extrapolation -- moving quickly to the prior mean, having learned
no structure to generalise beyond the data.  The SE kernel
interpolates somewhat sensibly, but appears to underestimate the
magnitudes of peaks and troughs, and treats repetitive patterns in the
data as noise.  Extrapolation using the SE kernel is poor.  The RQ
kernel, which is a scale mixture of SE kernels, is more able to manage
different length-scales in the data, and generalizes the long term
trend better than the SE kernel, but interpolates poorly.

By contrast, the SM kernel interpolates nicely (overlapping with the
data in the training region), and is able to extrapolate
complex patterns far beyond the data, capturing the true airline
passenger numbers for years after the data ends, within a small band
containing 95\% of the predictive probability mass.

Of the~${Q=10}$ initial components specified for the SM kernel,~$7$
were used after training.  The learned spectral density in
Figure \ref{fig: airlinefigures}b shows a large sharp low frequency
peak (at about $0.00148$).  This peak corresponds to the rising trend,
which generalises well beyond the data (small variance peak), is
smooth (low frequency), and is important for describing the data
(large relative weighting).  The next largest peak corresponds to the
yearly trend of 12 months, which again generalises, but not to the
extent of the smooth rising trend, since the variance of this peak is
larger than for the peak near the origin.  The higher frequency peak
at~${x=0.34}$ (period of 3 months) corresponds to the beginnings of
new seasons, which can explain the effect of seasonal holidays on air
traffic. 
 
A more detailed study of these features and their properties
-- frequencies, variances, etc. -- could isolate less
obvious features affecting airline passenger
numbers.  Table \ref{tab: predictions} (\texttt{AIRLINE}) shows 
predictive performance for forecasting 48 months of airline passenger numbers.

\begin{table}
\caption{We compare the test performance of the proposed spectral mixture (SM) kernel
with squared exponential (SE), Mat{\'e}rn (MA), rational quadratic (RQ), and periodic (PE) kernels. The SM kernel consistently has the lowest mean
squared error (MSE) and highest log likelihood ($\mathcal{L}$).}
\small
\begin{center}
\begin{tabular}{l r r r r r }
\toprule 
  &  SM & SE & MA & RQ & PE\\
\midrule             
 \texttt{CO}$_\texttt{2}$ &   & & & &     \\  
\midrule
  MSE & $\bm{9.5}$ & $1200$ & $1200$ & $980$ & $1200$ \\
  $\mathcal{L}$ & $\bm{170}$ & $-320$ & $-240$ & $-100$& $-1800$  \\
\midrule
  \texttt{NEG COV} &  &  & & &  \\  
\midrule
  MSE & $\bm{62}$ & $210$ & $210$& $210$& $210$ \\
  $\mathcal{L}$ & $\bm{-25}$ & $-70$ & $-70$ &$-70$ & $-70$ \\
\midrule
 \texttt{SINC} &  &  & & &   \\  
\midrule
  MSE & $\bm{0.000045}$ & $0.16$ & $0.10$ & $0.11$ & $0.05$   \\
  $\mathcal{L}$ & $\bm{3900}$ & $2000$ & $1600$ & $2000$ & $600$ \\
\midrule
 \texttt{AIRLINE} &  &  & & &   \\  
\midrule
  MSE & $\bm{460}$ & $43000$ & $37000$ & $4200$ & $46000$  \\
  $\mathcal{L}$ & $\bm{-190}$ & $-260$ & $-240$ & $-280$ & $-370$\\
\bottomrule
\end{tabular}
\end{center}
\label{tab: predictions}
\end{table}

\section{Discussion}

We have derived expressive closed form kernels and we have shown that
these kernels, when used with Gaussian processes, can discover
patterns in data and extrapolate over long ranges.  The simplicity of
these kernels is one of their strongest properties: they can be used
as drop-in replacements for popular kernels such as the squared
exponential kernel, with major benefits in expressiveness and
performance, while retaining simple training and inference 
procedures.

Gaussian processes have proven themselves as powerful smoothing
interpolators.  We believe that pattern discovery and extrapolation is
an exciting new direction for Bayesian nonparametric approaches, which 
can capture rich variations in data.  Here we have shown how Bayesian 
nonparametric models can naturally be used to generalise a pattern from 
a small number of examples.

We have only begun to explore what could be done
with such pattern discovery methods.  In future work, one could 
integrate away a spectral density, using recently developed
efficient Markov chain Monte Carlo for GP hyperparameters
\citep{murray-adams-2010a}.  Moreover, recent Toeplitz methods 
\citep{saatchi11} could be applied to the SM kernel for significant 
speedup in inference and predictions.

\textbf{Acknowledgements}
We thank Richard E.\ Turner, Carl Edward Rasmussen, David A.\ Knowles, and Neil D.\ Lawrence, for interesting discussions.

\nocite{wilson2011gaussian}

\bibliography{mbibnew}

\begin{thebibliography}{30}
\providecommand{\natexlab}[1]{#1}
\providecommand{\url}[1]{\texttt{#1}}
\expandafter\ifx\csname urlstyle\endcsname\relax
  \providecommand{\doi}[1]{doi: #1}\else
  \providecommand{\doi}{doi: \begingroup \urlstyle{rm}\Url}\fi

\bibitem[Abrahamsen(1997)]{abrahamsen1997}
Abrahamsen, P.
\newblock A review of {G}aussian random fields and correlation functions.
\newblock \emph{Norweigan Computing Center Technical report}, 1997.

\bibitem[Archambeau \& Bach(2011)Archambeau and Bach]{archambeau2011}
Archambeau, C. and Bach, F.
\newblock Multiple {G}aussian process models.
\newblock \emph{arXiv preprint arXiv:1110.5238}, 2011.

\bibitem[Bochner(1959)]{bochner1959lectures}
Bochner, Salomon.
\newblock \emph{Lectures on Fourier Integrals.(AM-42)}, volume~42.
\newblock Princeton University Press, 1959.

\bibitem[Chatfield(1989)]{chatfield1989}
Chatfield, C.
\newblock \emph{Time Series Analysis: An Introduction}.
\newblock London: Chapman and Hall, 1989.

\bibitem[Cressie(1993)]{cressie1993statistics}
Cressie, N.A.C.
\newblock \emph{Statistics for Spatial Data (Wiley Series in Probability and
  Statistics)}.
\newblock Wiley-Interscience, 1993.

\bibitem[Damianou \& Lawrence(2012)Damianou and Lawrence]{damianou2012deep}
Damianou, A.C. and Lawrence, N.D.
\newblock Deep {G}aussian processes.
\newblock \emph{arXiv preprint arXiv:1211.0358}, 2012.

\bibitem[Durrande et~al.(2011)Durrande, Ginsbourger, and
  Roustant]{durrande2011}
Durrande, N., Ginsbourger, D., and Roustant, O.
\newblock Additive kernels for {G}aussian process modeling.
\newblock \emph{arXiv preprint arXiv:1103.4023}, 2011.

\bibitem[G{\"o}nen \& Alpayd{\i}n(2011)G{\"o}nen and Alpayd{\i}n]{gonen2011}
G{\"o}nen, M. and Alpayd{\i}n, E.
\newblock Multiple kernel learning algorithms.
\newblock \emph{Journal of Machine Learning Research}, 12:\penalty0 2211--2268,
  2011.

\bibitem[Hyndman(2005)]{hyndman2005}
Hyndman, R.J.
\newblock Time series data library.
\newblock 2005.
\newblock \url{http://www-personal.buseco.monash.edu.au/~hyndman/TSDL/}.

\bibitem[Keeling \& Whorf(2004)Keeling and Whorf]{keeling2004}
Keeling, C.~D. and Whorf, T.~P.
\newblock Atmospheric {CO2} records from sites in the {SIO} air sampling
  network.
\newblock \emph{Trends: A Compendium of Data on Global Change. Carbon Dioxide
  Information Analysis Center}, 2004.

\bibitem[Kostantinos(2000)]{kostantinos2000}
Kostantinos, N.
\newblock Gaussian mixtures and their applications to signal processing.
\newblock \emph{Advanced Signal Processing Handbook: Theory and Implementation
  for Radar, Sonar, and Medical Imaging Real Time Systems}, 2000.

\bibitem[MacKay(1998)]{mackay98}
MacKay, David~J.C.
\newblock Introduction to {G}aussian processes.
\newblock In Christopher M.~Bishop, editor (ed.), \emph{Neural Networks and
  Machine Learning}, chapter~11, pp.\  133--165. Springer-Verlag, 1998.

\bibitem[MacKay et~al.(1994)]{mackay1994}
MacKay, D.J.C. et~al.
\newblock Bayesian nonlinear modeling for the prediction competition.
\newblock \emph{Ashrae Transactions}, 100\penalty0 (2):\penalty0 1053--1062,
  1994.

\bibitem[McCulloch \& Pitts(1943)McCulloch and Pitts]{mcculloch1943}
McCulloch, W.S. and Pitts, W.
\newblock A logical calculus of the ideas immanent in nervous activity.
\newblock \emph{Bulletin of mathematical biology}, 5\penalty0 (4):\penalty0
  115--133, 1943.

\bibitem[Murray \& Adams(2010)Murray and Adams]{murray-adams-2010a}
Murray, Iain and Adams, Ryan~P.
\newblock Slice sampling covariance hyperparameters in latent {G}aussian
  models.
\newblock In \emph{Advances in Neural Information Processing Systems 23}, 2010.

\bibitem[Neal(1996)]{neal1996}
Neal, R.M.
\newblock \emph{{Bayesian Learning for Neural Networks}}.
\newblock Springer Verlag, 1996.
\newblock ISBN 0387947248.

\bibitem[Rasmussen(1996)]{rasmussenphd96}
Rasmussen, Carl~Edward.
\newblock \emph{Evaluation of Gaussian Processes and Other Methods for
  Non-linear Regression}.
\newblock PhD thesis, University of Toronto, 1996.

\bibitem[Rasmussen \& Williams(2006)Rasmussen and Williams]{rasmussen06}
Rasmussen, Carl~Edward and Williams, Christopher~K.I.
\newblock \emph{Gaussian processes for Machine Learning}.
\newblock The {MIT} {P}ress, 2006.

\bibitem[Rosenblatt(1962)]{rosenblatt1962}
Rosenblatt, F.
\newblock \emph{Principles of Neurodynamics}.
\newblock Spartan Book, 1962.

\bibitem[Rumelhart et~al.(1986)Rumelhart, Hinton, and
  Williams]{rumelhart1986learning}
Rumelhart, D.E., Hinton, G.E., and Williams, R.J.
\newblock Learning representations by back-propagating errors.
\newblock \emph{Nature}, 323\penalty0 (6088):\penalty0 533--536, 1986.

\bibitem[Saatchi(2011)]{saatchi11}
Saatchi, Yunus.
\newblock \emph{Scalable Inference for Structured Gaussian Process Models}.
\newblock PhD thesis, University of Cambridge, 2011.

\bibitem[Salakhutdinov \& Hinton(2008)Salakhutdinov and
  Hinton]{salakhutdinov2008}
Salakhutdinov, R. and Hinton, G.
\newblock Using deep belief nets to learn covariance kernels for {G}aussian
  processes.
\newblock \emph{Advances in {N}eural {I}nformation {P}rocessing {S}ystems},
  20:\penalty0 1249--1256, 2008.

\bibitem[Stein(1999)]{stein1999}
Stein, M.L.
\newblock \emph{Interpolation of Spatial Data: Some Theory for Kriging}.
\newblock Springer Verlag, 1999.

\bibitem[Steyvers et~al.(2006)Steyvers, Griffiths, and Dennis]{steyvers2006}
Steyvers, M., Griffiths, T.L., and Dennis, S.
\newblock Probabilistic inference in human semantic memory.
\newblock \emph{Trends in Cognitive Sciences}, 10\penalty0 (7):\penalty0
  327--334, 2006.

\bibitem[Tenenbaum et~al.(2011)Tenenbaum, Kemp, Griffiths, and
  Goodman]{tenenbaum2011}
Tenenbaum, J.B., Kemp, C., Griffiths, T.L., and Goodman, N.D.
\newblock How to grow a mind: Statistics, structure, and abstraction.
\newblock \emph{Science}, 331\penalty0 (6022):\penalty0 1279--1285, 2011.

\bibitem[Tipping(2004)]{tipping2004}
Tipping, M.
\newblock Bayesian inference: An introduction to principles and practice in
  machine learning.
\newblock \emph{Advanced {L}ectures on {M}achine {L}earning}, pp.\  41--62,
  2004.

\bibitem[Wilson \& Adams(2013)Wilson and Adams]{wilsonsuppnew}
Wilson, A.G. and Adams, R.P.
\newblock Gaussian process kernels for pattern discovery and extrapolation
  supplementary material and code.
\newblock 2013.
\newblock \url{http://mlg.eng.cam.ac.uk/andrew/smkernelsupp.pdf}.

\bibitem[Wilson et~al.(2012)Wilson, Knowles, and Ghahramani]{wilson12icml}
Wilson, Andrew~G., Knowles, David~A., and Ghahramani, Zoubin.
\newblock Gaussian process regression networks.
\newblock In \emph{Proceedings of the 29th International Conference on Machine
  Learning (ICML)}, 2012.

\bibitem[Wilson et~al.(2011)Wilson, Knowles, and
  Ghahramani]{wilson2011gaussian}
Wilson, Andrew~Gordon, Knowles, David~A, and Ghahramani, Zoubin.
\newblock Gaussian process regression networks.
\newblock \emph{arXiv preprint arXiv:1110.4411}, 2011.

\bibitem[Yuille \& Kersten(2006)Yuille and Kersten]{yuille2006vision}
Yuille, A. and Kersten, D.
\newblock Vision as {B}ayesian inference: analysis by synthesis?
\newblock \emph{Trends in {C}ognitive {S}ciences}, 10\penalty0 (7):\penalty0
  301--308, 2006.

\end{thebibliography}
\bibliographystyle{icml2013}

\end{document}